\documentclass[sigconf]{acmart}

\usepackage{multirow}
\usepackage{makecell}
\usepackage[linesnumbered,ruled,vlined]{algorithm2e}
\usepackage{enumitem}
\usepackage{amsmath}
\usepackage{makecell}
\usepackage{subfigure}
\usepackage{etoolbox}
\SetKwInput{KwInput}{Input}                
\SetKwInput{KwOutput}{Output}              
\SetKwInput{KwTraining}{Training} 
\SetKwFor{KwFor}{For} 
\usepackage{balance} 

\AtBeginDocument{%
  }

\copyrightyear{2024}
\acmYear{2024}
\setcopyright{acmlicensed}\acmConference[KDD '24]{Proceedings of the 30th ACM SIGKDD Conference on Knowledge Discovery and Data Mining}{August 25--29, 2024}{Barcelona, Spain}
\acmBooktitle{Proceedings of the 30th ACM SIGKDD Conference on Knowledge Discovery and Data Mining (KDD '24), August 25--29, 2024, Barcelona, Spain}
\acmDOI{10.1145/3637528.3671974}
\acmISBN{979-8-4007-0490-1/24/08}

\begin{document}

\title{FlexCare: Leveraging Cross-Task Synergy for Flexible Multimodal Healthcare Prediction}


\author{Muhao Xu}
\affiliation{%
  \institution{Institute of Information Science, Beijing Jiaotong University}
  \institution{Beijing Key Laboratory of Advanced Information Science and Network Technology}
  \city{Beijing}
  \country{China}
}
\email{mhxu1998@bjtu.edu.cn}

\author{Zhenfeng Zhu}
\authornote{Corresponding author.}
\affiliation{%
  \institution{Institute of Information Science, Beijing Jiaotong University}
  \institution{Beijing Key Laboratory of Advanced Information Science and Network Technology}
  \city{Beijing}
  \country{China}
}
\email{zhfzhu@bjtu.edu.cn}

\author{Youru Li}
\affiliation{
  \institution{Institute of Information Science, Beijing Jiaotong University}
  \institution{Beijing Key Laboratory of Advanced Information Science and Network Technology}
  \city{Beijing}
  \country{China}
}
\email{liyouru@bjtu.edu.cn}

\author{Shuai Zheng}
\affiliation{
  \institution{Institute of Information Science, Beijing Jiaotong University}
  \institution{Beijing Key Laboratory of Advanced Information Science and Network Technology}
  \city{Beijing}
  \country{China}
}
\email{zs1997@bjtu.edu.cn}

\author{Yawei Zhao}
\author{Kunlun He}
\affiliation{
  \institution{Medical Big Data Research Center, Chinese PLA General Hospital}
  \city{Beijing}
  \country{China}
}
\email{csyawei.zhao@gmail.com}
\email{kunlunhe@plagh.org}

\author{Yao Zhao}
\affiliation{
  \institution{Institute of Information Science, Beijing Jiaotong University}
  \institution{Beijing Key Laboratory of Advanced Information Science and Network Technology}
  \city{Beijing}
  \country{China}
}
\email{yzhao@bjtu.edu.cn}

\renewcommand{\shortauthors}{Muhao Xu et al.}

\begin{abstract}
  Multimodal electronic health record (EHR) data can offer a holistic assessment of a patient's health status, supporting various predictive healthcare tasks.
  Recently, several studies have embraced the multitask learning approach in the healthcare domain, exploiting the inherent correlations among clinical tasks to predict multiple outcomes simultaneously.
  However, existing methods necessitate samples to possess complete labels for all tasks, which places heavy demands on the data and restricts the flexibility of the model.
  Meanwhile, within a multitask framework with multimodal inputs, 
  how to comprehensively consider the information disparity among modalities and among tasks still remains a challenging problem.
  To tackle these issues, a unified healthcare prediction model, also named by \textbf{FlexCare}, is proposed to flexibly accommodate incomplete multimodal inputs, promoting the adaption to multiple healthcare tasks. 
  The proposed model breaks the conventional paradigm of parallel multitask prediction by decomposing it into a series of asynchronous single-task prediction.
  Specifically, a task-agnostic multimodal information extraction module is presented to capture decorrelated representations of diverse intra- and inter-modality patterns.
  Taking full account of the information disparities between different modalities and different tasks,
  we present a task-guided hierarchical multimodal fusion module that integrates the refined modality-level representations into an individual patient-level representation.
  Experimental results on multiple tasks from MIMIC-IV/MIMIC-CXR/MIMIC-NOTE datasets demonstrate the effectiveness of the proposed method.
  Additionally, further analysis underscores the feasibility and potential of employing such a multitask strategy in the healthcare domain. 
  The source code is available at \href{https://github.com/mhxu1998/FlexCare}{https://github.com/mhxu1998/FlexCare}.
\end{abstract}




\maketitle

\section{Introduction}

\begin{figure}[t]
  \centering
  \includegraphics[width=0.95\linewidth]{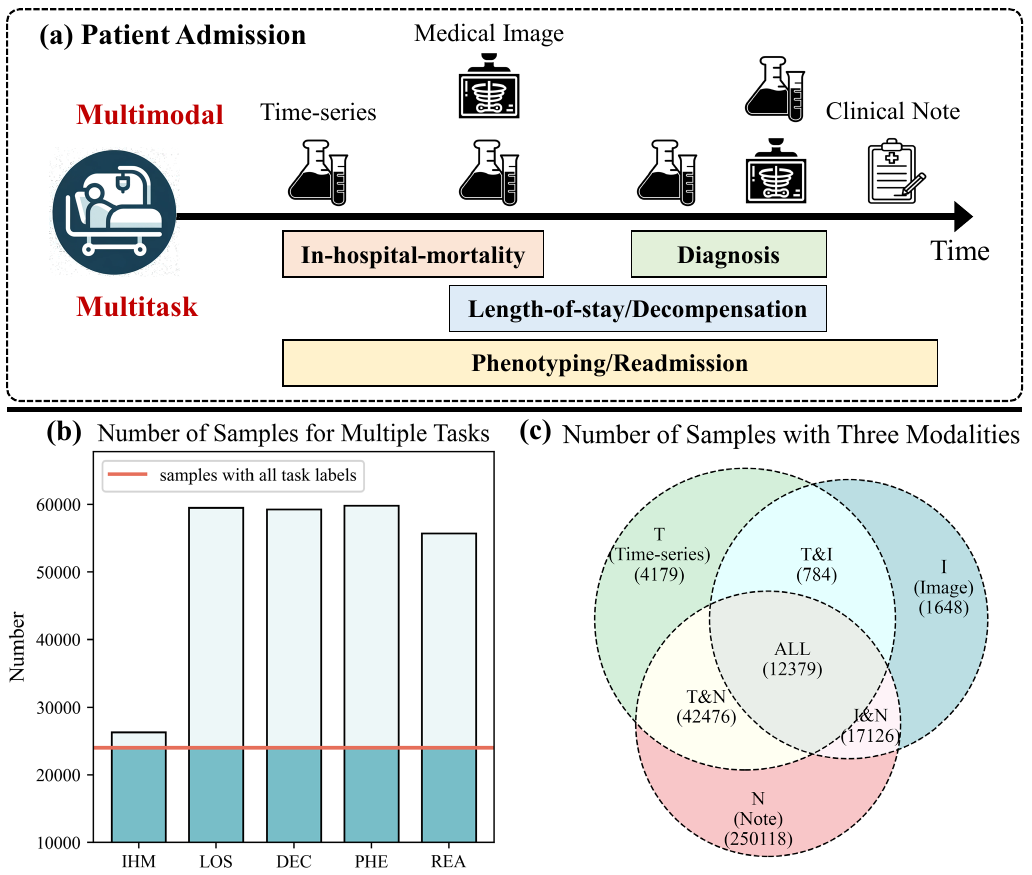}
  \caption{(a) Illustration of the multimodal data and the multitask predictions during a patient's admission; (b) The number of samples from multiple tasks that depend on time-series data in the MIMIC-IV dataset; (c) The number of samples with different modality data in the MIMIC-IV dataset. }
  \label{fig:intro}
\end{figure}

To comprehensively assess a patient's health status, clinical practice often employs a variety of methods to capture diverse patient information, resulting in multimodal EHR data that comprise both structured and unstructured data. 
These diverse multimodal EHR data can underpin multiple predictive tasks in the clinical, facilitating the identification of high-risk patients for early intervention. For instance, time-series data such as vital signs and laboratory test results are frequently utilized for clinical risk and outcome prediction~\cite{ma2020concare,XuYQZZWWX23}. 
Medical images (e.g., X-rays and computerized tomography scans), are instrumental in detecting, localizing, and classifying diseases relevant to patients~\cite{irvin2019chexpert}.
Compared to structured information, textual clinical notes furnish comprehensive insights into a patient's medical history, symptoms, and the reasoning for diagnoses, offering a more macroscopic and holistic perspective~\cite{li2020icd,rasmy2021med}.
Considering the complementary nature of information within multimodal EHR data, some previous methodologies have been developed to integrate various modalities for enhancing the accuracy of clinical event prediction~\cite{soenksen2022integrated,hayat2022medfuse,zhang2022m3care,pellegrini2023unsupervised,lee2023learning}.
Furthermore, some  studies~\cite{harutyunyan2019multitask,aoki2022heterogeneous,zhao2022unimed} leverage the inherent correlations among clinical tasks, employing multitask learning approaches to simultaneously predict multiple tasks.

However, current multitask models for healthcare prediction typically necessitate complete labels for all tasks~\cite{harutyunyan2019multitask,zhao2022unimed}. Such demand for data is exceedingly stringent, especially within the medical domain.
Indeed, the requirements for EHR data manifest considerable variation across different clinical tasks for the same patient, encompassing distinct time spans and modalities, as illustrated in Figure \ref{fig:intro}(a).
For example, in-hospital-mortality prediction commonly relies on time-series data from the first 48 hours after a patient is admitted to the hospital, while disease diagnosis mandates the presence of image modality, with no restrictions on the other modalities.
Hence, requiring identical input data and complete labeling for multitask constitutes a significant waste of already scarce healthcare data. 
The statistical information for the MIMIC-IV dataset~\cite{johnson2020mimic,johnson2019mimic,johnson2023mimic} presented in Figure \ref{fig:intro}(b) and (c) illustrates that samples with all task labels and samples with all modalities represent only a small fraction of the total dataset.
In summary, to overcome the limitations of previous models that require completeness in data and labels, multimodal multitask healthcare prediction models face the following challenges:

\begin{figure}[t]
  \centering
  \includegraphics[width=1\linewidth]{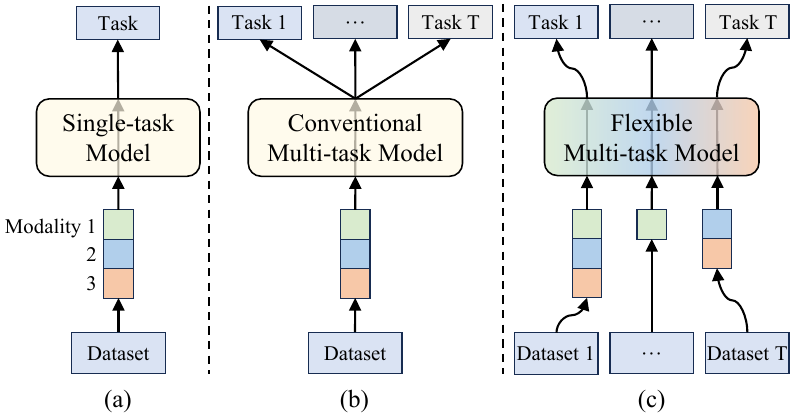}
  \caption{(a) Single-task model; (b) Conventional multi-task model; (c) Our proposed flexible multi-task model.}
  \label{fig:intro2}
\end{figure}

\textbf{Challenge 1: How to develop a flexible model capable of supporting multimodal inputs and adapting to various heterogeneous tasks, without requiring comprehensive labels for each sample across all tasks?}
\textit{Flexibility} here is manifested in the model not requiring each sample to possess inputs for all modalities and labels for all tasks. 
An intuitive approach is to deconstruct parallel multitask simultaneous predictions into asynchronous multiple single-task predictions, where each sample corresponds to a single task label, thus satisfying the data differences in terms of time spans and modalities used between multiple tasks for the same patient.
As shown in Figure \ref{fig:intro2}, the hallmark of a flexible multitask model lies in employing a unified model to process multiple heterogeneous datasets, encompassing both heterogeneous multimodal inputs and heterogeneous tasks.
In the realm of universal natural language and visual understanding, MT-DNN~\cite{liu2019multi} and Unit~\cite{hu2021unit} have made attempts with a multitask approach using heterogeneous datasets.
However, current studies~\cite{harutyunyan2019multitask,aoki2022heterogeneous,zhao2022unimed} still lack such consideration in the field of healthcare prediction.

\textbf{Challenge 2: How to deal with the information disparities among modalities and tasks comprehensively within a multitask framework?}
Three forms of information disparity need to be considered in the unified framework: modality-modality disparity, task-task disparity, and task-data disparity. 
Concretely, in the heterogeneous EHR data, different modalities may encapsulate distinct aspects of a patient's health status. It is imperative that this informational disparity present among multimodal data be taken into consideration.
Furthermore, different clinical tasks focus on EHR data from various time spans and modalities of the patient, posing challenges to the construction of the multitask model.
On the other hand, due to the extensive sharing of modules in multitask models, negative interference may arise when inter-task correlations are weak.
In multimodal multitask models, the shared modules are not only cross-task but also cross-modal, further complicating the above issue.

By jointly considering the above issues, we propose a model leveraging cross-task synergy for \textbf{flex}ible multimodal health\textbf{care} prediction (\textbf{FlexCare}).
Our main contributions are summarized as follows:
\begin{itemize}[leftmargin=*,noitemsep,topsep=2pt]
    \item To the best of our knowledge, this work is the first attempt to study a unified healthcare prediction model that flexibly supports multimodal inputs and adapts to healthcare multitask. It is empowered with multitasking capabilities in the form of multiple single-task asynchronous predictions.
    \item To comprehensively capture the information from various modality combination patterns, a task-agnostic multimodal information extraction module is presented, with the covariance regularization to decorrelate the different modality combination representations.
    \item A task-guided hierarchical multimodal fusion module is designed to learn adaptive representations for different tasks, by incorporating implicit task indication in the aggregation process from modality-level to patient-level representation.
    \item Extensive experiments conducted on MIMIC-IV dataset show that our model achieves competitive results on multiple tasks. Besides, further analysis demonstrates the feasibility and potential of adopting such multitask strategy to construct a unified model in the healthcare domain.
\end{itemize}

\section{Related work}
\noindent\textbf{Multimodal learning for healthcare.} 
In the quest for a thorough comprehension of patient patterns to enhance the precision of clinical event prediction, researchers have delved into the realm of multimodal learning utilizing healthcare data~\cite{zhang2021learning,golovanevsky2022multimodal,zheng2022multi,liu2022multimodal,tang2023predicting,pellegrini2023unsupervised,zhang2023improving,xu2023vecocare}.
HAIM~\cite{soenksen2022integrated} leverages different pre-trained feature extraction models to process multimodal inputs and obtains the overall representation of the patient.
Zhou et al.~\cite{zhou2023transformer} proposes a transformer-based model that processes multimodal EHR data in a unified manner.
However, the aforementioned methods lack consideration for incomplete modalities, which limits the application scenarios of the models.

To handle the pervasive issue of missing modalities in clinical practice, many researchers have developed models capable of either imputing missing modalities or adapting to the absence of certain modalities.
MedFuse~\cite{hayat2022medfuse} is an LSTM-based fusion model that can process uni-modal as well as multimodal input.
Lee et al.~\cite{lee2023learning} learns the EHR data with missing modal by the modality-aware attention with skip bottleneck.
To overcome the limitations of traditional parallel fusion approaches, MultiModN~\cite{swamy2023multimodn} sequentially inputs any number or combination of modalities into a sequence of modality-specific encoders and can skip over missing modalities.
Compared with the above method of ignoring missing modalities, M3Care~\cite{zhang2022m3care} imputes the information of the missing modalities in the latent space from the similar neighbors of each patient.
While these methods are effective, when extending to a multitasking setting, the process of handling multimodal information must take into account the different focal points of various tasks.

\noindent\textbf{Multitask learning for healthcare.} 
Multitask learning, an effective method that enhances performance through the joint learning of multiple related tasks, has been explored for use in healthcare prediction.
In the healthcare domain, the term "tasks" has different definitions.
Several studies \cite{suresh2018learning,liu2020multi} treat the mortality prediction of different patient cohorts as multiple tasks.
Some studies introduce auxiliary tasks, such as time series reconstruction \cite{yu2019using} and prediction based on unimodal data \cite{xu2024redcdr}, to enhance the model's representational capacity and downstream task performance.
Other studies predict multiple clinical tasks simultaneously, such as risk prediction~\cite{harutyunyan2019multitask,si2019deep,aoki2022heterogeneous,xu2023cooperative} and disease diagnosis~\cite{kyono2019multi,shao2020multi}.
GenHPF \cite{hur2023genhpf} converts EHRs into hierarchical textual representations and offers a solid framework for multi-task and multi-source learning, but it can only handle EHRs in text form.
Recently, UniMed~\cite{zhao2022unimed} sequentially predict four medical tasks based on multimodal EHR data, utilizing the time-progressive correlation between tasks.
However, these methods require samples to have labels for all tasks, imposing stricter demands on the already scarce medical data.

\begin{table}[b]
    \centering\caption{Notations used in this paper}
    \label{tab:notations}{
    \begin{tabular}{c|p{6.5cm}}
        \hline
        Notation & Definition \\
        \hline
        $\tau$; $T$ & The $\tau$-th task; total number of tasks\\
        $\mathcal{D}_{\tau}$; $N_{\tau}$ & The dataset of the $\tau$-th task; the number of samples \\
        $m$; $\mathcal{M}$ & Modality; set of multimodal\\
        $c$; $\mathcal{C}$ & Modality combination; set of modality combination\\
        $\mathbf{X}_{\tau}^{(n)}$  & The $n$-th patient's data of the $\tau$-th task\\
        $y_{\tau}^{(n)}$ & Ground truth of the target\\
        $\hat{y}_{\tau}^{(n)}$ & Prediction result\\
        \hline        
        $\mathbf{H}_{\tau}^m$  & Learned representations of modality $m$\\
        $\mathbf{h}_{\tau}^\text{task}$; $\mathbf{H}_{\tau}^\text{comb}$ &  Learnable representation of task token and modality combination tokens  \\

        $\mathbf{z}_{\tau}^\text{task}$; $\mathbf{Z}_{\tau}^{\text{comb}}$ & Learned representation of task token and modality combination tokens through intra/inter-modality encoder  \\
        $\mathbf{M}_{\tau}$ & The mask that aids modality combination tokens in capturing specific information \\
        $C_{\tau}$    & The covariance regularization term of $\mathbf{Z}_{\tau}^{\text{comb}}$ \\
         \hline
        $\mathbf{s}_{\tau}^c$ & The refined representation of modality combination $c$\\
        $\mathbf{s}_{\tau}^p$  & The patient-level representation of the patient\\

         \hline
    \end{tabular}}
\end{table}

\begin{figure*}[t]
  \centering
  \includegraphics[width=1\linewidth]{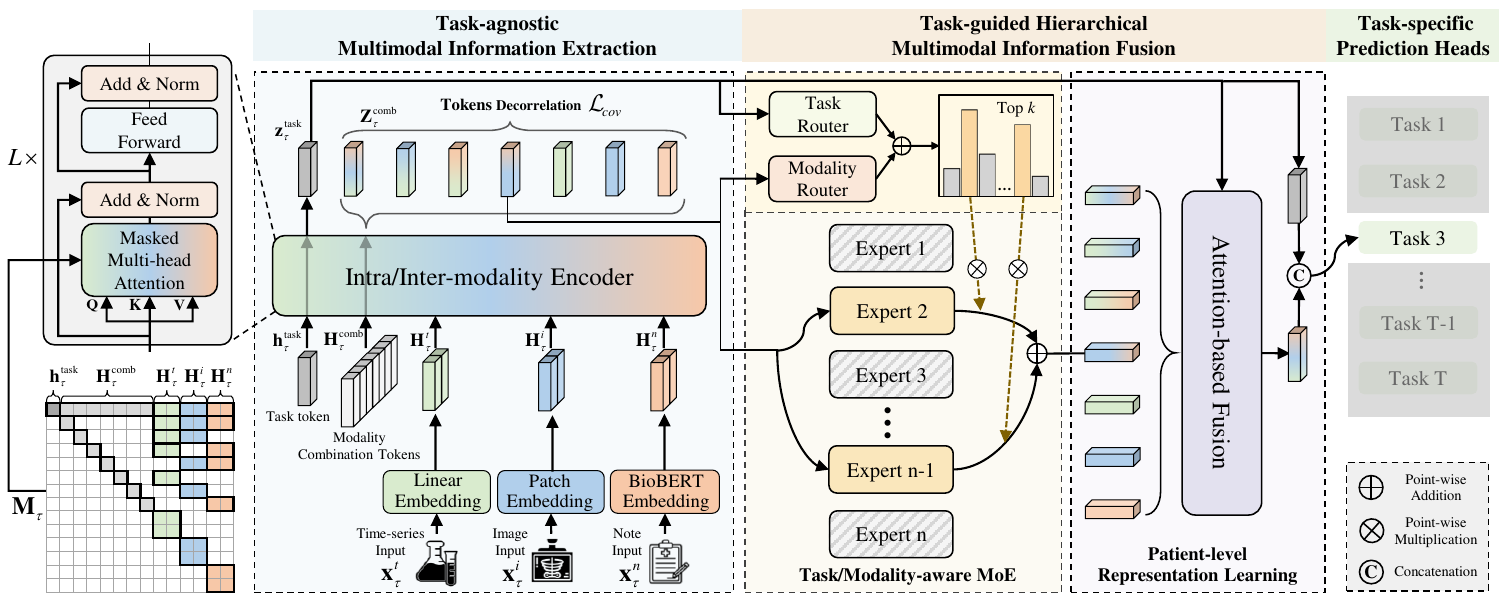}
  \caption{The framework of the FlexCare model. It consists of three modules: (a) Task-agnostic multimodal information extraction; (b) Task-guided hierarchical multimodal fusion; (c) Task-specific prediction heads.}
  \label{fig:framework}
\end{figure*}

\section{Problem Formulation}
In this section, we present the problem formulation and symbol notation used throughout the paper. 

\noindent\textbf{Definition 1 (Patient multitask data).}
The datasets for a set of tasks are denoted as $\{\mathcal{D}_{\tau}\}_{\tau=1}^{T}$, where $T$ is the number of tasks. Thus, the corresponding training set of the $\tau$-th task is represented as: $\mathcal{D}_{\tau} = \left\{ (\mathbf{X}_{\tau}^{(n)}, y_{\tau}^{(n)}) \right\}_{n=1}^{N_{\tau}}$, where $N_{\tau}$ is the number of samples, $\mathbf{X}_{\tau}^{(n)}$ and $y_{\tau}^{(n)} \in \mathcal{Y}_{\tau}$ are the multimodal input and ground truth of the $n$-th sample, respectively. $\mathcal{Y}_{\tau}$ is the set of labels for the $\tau$-th task.

\noindent\textbf{Definition 2 (Patient multimodal data).}
Given $\mathcal{M}=\{t,i,n\}$ a set of modalities (i.e., time-series data, image, note), the input of $n$-th sample can be defined as: $\mathbf{X}_{\tau}^{(n)}=\{\mathbf{X}_{\tau}^{(n),m}\}_{m\in \mathcal{M}}$.
Considering the absence of some modalities, the incomplete input is $\mathbf{X}_{\tau}^{(n)}=\{\mathbf{X}_{\tau}^{(n),m}\}_{m\in \mathcal{M}^{(n)}}$, where $\mathcal{M}^{(n)}$ are the modalities actually present in the $n$-th sample, and $\mathcal{M}^{(n)}\subseteq \mathcal{M}$. Note that $|\mathcal{M}^{(n)}| \geq 1$, because at least one modality is present for each sample.

\noindent\textbf{Definition 3 (Modality combination).}
The modality combination set represents all patterns of unimodal or multimodal combination, defined as: $\mathcal{C}=2^\mathcal{M}\textbackslash \emptyset$ (i.e., all nonempty subsets of $\mathcal{M}$).
When $|\mathcal{M}|=3$, the number of modality combination set $|\mathcal{C}|=7$.

\noindent\textbf{Multitask prediction problem.}
Given the multimodal datasets for different tasks, the objective is to learn a unified task-adaptive predictive function: $\hat{y}_{\tau}^{(n)}=f_{\theta}(\mathbf{X}_{\tau}^{(n)},\tau)$, where $f_{\theta}(\cdot,\cdot)$ is parameterized by $\theta$ and $\hat{y}_{\tau}^{(n)}$ denotes the corresponding prediction result of the $\tau$-th task.

Besides, the necessary notations used in the paper are listed in Table~\ref{tab:notations} for ease of understanding. Please note that in the Section 4 Methodology, we take a single patient as an example. Therefore, we simplify $\mathbf{X}_{\tau}^{(n)}$ to $\mathbf{X}_{\tau}$ to enhance the reading experience.

\section{Methodology}

\subsection{Overview}
The overall framework of the proposed model is shown in Figure \ref{fig:framework}, which mainly consists of three components:
\begin{itemize}[leftmargin=*,noitemsep,topsep=2pt]
    \item The \textbf{Task-agnostic multimodal information extraction} module leverages unimodal feature extractors and a unified multimodal encoder to learn a spectrum of modality combination representations.
    \item The \textbf{Task-guided hierarchical multimodal fusion} module achieves hierarchical fusion from modality-level to patient-level through the task/modality-aware Mixture of Experts (MoE) and an attention-based fusion mechanism.
    \item The \textbf{Task-specific prediction heads} are configured with individual predictors for each task, making predictions for the current task based on patient-level representation.
\end{itemize}
Due to the model being trained and tested in an asynchronous multiple single-task paradigm, the following introduction will take one task as an example.

\subsection{Task-agnostic Multimodal Information Extraction}
To enhance the generalization capability of the model, the task-agnostic multimodal information extraction module is deeply shared across multiple tasks. The module maps raw data with different dimensions and modalities into latent representations in a unified space, and captures task information and modality-level information through learnable task token and modality combination tokens respectively.
\subsubsection{\textbf{Unimodal Information Extraction}}

Given the sample $\mathbf{X}_{\tau}$ in the $\tau$-th task dataset and the unimodal representation extraction model $f_m(\cdot)$, the raw input of modality $m$ is converted into a series of 1D tokens $\mathbf{H}^m_{\tau}\in \mathbb{R}^{N^m\times d}$ with the same dimension via:
\begin{equation}
\mathbf{H}^m_{\tau}=f_m(\mathbf{X}^m_{\tau})+\mathbf{p}^m,
\label{eq:unimodal}
\end{equation}
where $\mathbf{p}^m\in \mathbb{R}^{N^m\times d}$
is a learned positional embedding added to the tokens to retain positional information, $d$ is the dimension of the latent representation and $N^m$ denotes the number of tokens for modality $m$. 
For the three modalities $\{t,i,n\}$, the unimodal models $f_m(\cdot)$ are set as follows: a linear projection, patch projection~\cite{DosovitskiyB0WZ21} followed by a linear projection, and the pre-trained \& frozen BioBERT~\cite{lee2020biobert}, respectively.

\subsubsection{\textbf{Intra/Inter-modality Encoder}}

To inject task-specific information into the model, thereby facilitating subsequent implicit guidance for multimodal representation learning under specific tasks, a learnable task token is allocated for each task category. For the given $\tau$-th task, its corresponding task token is represented as $\mathbf{h}^{\text{task}}_{\tau}\in \mathbb{R}^d$.
Furthermore, in order to thoroughly capture the information contained in individual intra-modality as well as various inter-modality patterns, a set of learnable modality combination tokens $\mathbf{H}^{\text{comb}}_{\tau}=\{\mathbf{h}^c_{\tau}\}_{c\in\mathcal{C}}$ is presented. 

To realize the information extraction from intra- and inter-modality, the task token, modality combination tokens, and tokens extracted from each modality, are stacked along the token dimension to obtain a multimodal sequence:
$\mathbf{H}_{\tau}^{0}=[\mathbf{h}_{\tau}^{\text{task}},\mathbf{H}_{\tau}^{\text{comb}},\mathbf{H}_{\tau}^{t},\mathbf{H}_{\tau}^{i},\mathbf{H}_{\tau}^{n}]$, where $\mathbf{H}_{\tau}^{0}\in \mathbb{R}^{N^h\times d}$ and $N^{h}=1+|\mathcal{C}|+N^t+N^i+N^n$.
Subsequently, the multimodal sequence is fed into the encoder, facilitating the interaction of multimodal information and obtaining the output of multimodal fusion:
\begin{equation}
\begin{split}
    \widetilde{\mathbf{H}}_{\tau}^l&=\text{LN}(\mathbf{H}_{\tau}^{l-1}+\text{M-MHSA}(\mathbf{H}_{\tau}^{l-1},\mathbf{H}_{\tau}^{l-1},\mathbf{H}_{\tau}^{l-1},\mathbf{M}_{\tau})),\\
    \mathbf{H}_{\tau}^l&=\text{LN}(\widetilde{\mathbf{H}}_{\tau}^l+\text{FFN}(\widetilde{\mathbf{H}}_{\tau}^l)),\\
\end{split}\label{eq:encoder}
\end{equation}
where $l=1,..., L$ is the number of encoder layers. $\text{FFN}(\cdot)$ and $\text{LN}(\cdot)$ refer to the feed-forward network and layer normalization, respectively.
$\text{M-MHSA}(\cdot,\cdot,\cdot,\cdot)$ represents a specially designed masked multi-head self-attention mechanism, defined as follows:
\begin{equation}
    \text{M-MHSA}(\mathbf{Q},\mathbf{K},\mathbf{V},\mathbf{M})=\text{Softmax}(\frac{\mathbf{Q}\mathbf{W}^Q(\mathbf{K}\mathbf{W}^K)^\mathsf{T}}{\sqrt{d}}+\mathbf{M})\mathbf{V}\mathbf{W}^V,
\label{eq:mhsa}
\end{equation}
where $\mathbf{Q},\mathbf{K},\mathbf{V}$ represent the query, key, and value embedding, respectively. $\mathbf{W}^Q,\mathbf{W}^K,\mathbf{W}^V$ denote the weight matrices, $d$ is the dimension of the latent representation, and $\mathbf{M}$ is the additional mask.

The additional mask $\mathbf{M}_{\tau}\in \mathbb{R}^{N^h\times N^h}$ enables modality combination tokens to precisely target information relevant to diverse modality combination patterns. Specifically,
for modality combination tokens and modality tokens, (i.e., $0<i,j<N^h$), the mask is denoted as:
\begin{equation}
    \mathbf{M}_{\tau}(i,j)=\left\{\begin{aligned}
        &0, \quad  \phi(j)\in\phi(i) \text{ or } \phi(j)=\phi(i) \\ 
        &-\infty, \quad \text{otherwise}
\end{aligned},\right.
\end{equation}
where $\phi:\text{index}\mapsto(\mathcal{M} |\mathcal{C})$ defines a function that maps the token to the modality $m$ or modality combination $c$ it belongs to.
$\phi(j)\in \phi(i)$ indicates that the token $\mathbf{H}_{\tau,j}$ originates from modality $m$, the token $\mathbf{H}_{\tau,i}=\mathbf{h}_{\tau}^c$ from modality combination $c$, and $m\in c$.
$\phi(j)=\phi(i)$ indicates that the token $\mathbf{H}_{\tau,j}$ and token $\mathbf{H}_{\tau,i}$ is from the same modality $m$ or modality combination $c$.

Given our intention for the multimodal information extraction process to be task-agnostic, the task token exclusively aggregates information from other tokens unidirectionally, without transmitting its own information to them.
Consequently, for the mask associated with the task token (i.e., $i=0$ and $0\leq j <N^h$), we establish that $\mathbf{M}_{\tau}(i,j)=0$.

Through $L$ layers of the encoder equipped with a specialized attention mask, we have obtained representation enriched with task-specific information $\mathbf{z}_{\tau}^\text{task}=\mathbf{h}_{\tau}^{\text{task},L}$ and various task-agnostic modality combination representations $\mathbf{Z_{\tau}^\text{comb}}=\mathbf{H}_{\tau}^{\text{comb},L}$.

\subsubsection{\textbf{Representation Decorrelation of Modality Combination}}
The modality combination tokens, due to the sharing of partial modality information within the encoder, may result in a phenomenon of feature redundancy.
To decorrelate the different embeddings of the modality combination tokens and prevent them from encoding similar information, a token-level covariance regularization method is proposed.

Unlike the previous method~\cite{BardesPL22} that constrains the correlations between different dimensions of the embeddings, our approach computes the covariance matrix along the token dimension rather than the feature dimension.
The function to calculate the covariance matrix $\text{Cov}(\cdot)$ and the token-level covariance regularization term $C_{\tau}$ are defined as follows: 
\begin{equation}
    \text{Cov}(\mathbf{Z})=\frac{1}{d-1}\sum_{j=1}^d(\mathbf{z}_{:,j}-\overline{\mathbf{z}})(\mathbf{z}_{:,j}-\overline{\mathbf{z}})^\mathsf{T},\text{where}\quad \overline{\mathbf{z}} = \sum_{j=1}^d\mathbf{z}_{:,j},
\end{equation}
\begin{equation}
    C_{\tau}=\frac{1}{(|\mathcal{C}|-1)^2}\sum_{i\neq j}[\text{Cov}(\mathbf{Z}_{\tau}^{\text{comb}})]_{i,j}^{2}.
\end{equation}

The loss function encourages the off-diagonal elements of the covariance matrix to approach 0, compelling the modality combination tokens to capture multimodal information that is uncorrelated with one another:
\begin{equation}
    \mathcal{L}_{cov}=\frac{1}{N_{\tau}}\sum_{n=1}^{N_{\tau}}C_{\tau}^{(n)}
\end{equation}

\subsection{Task-guided Hierarchical Multimodal Fusion}
To facilitate the adaptive representation learning for the specific task, it is necessary to infuse implicit task information into task-agnostic modality-level representations for their refinement. Guided by the specific task, these representations are subsequently aggregated into patient-level representations.
\subsubsection{\textbf{Task/Modality-aware Mixture of Experts}}
Within a multitask multimodal framework, it is imperative to consider the information disparities among modalities and tasks. Employing identical network layers for different modalities across varied tasks may lead to negative interference, hindering the individualized representation learning for specific downstream tasks.

Specifically, after acquiring the task token and various modality combination tokens from the original embeddings, a task/modality-aware Mixture-of-Experts module is employed for the refinement of multimodal representations in the context of a specific task. 
The module obtains the output $\mathbf{s}_{\tau}^c\in \mathbb{R}^{d}$ for the input token $\mathbf{z}_{\tau}^c$ by weighted average of the selected $k$ experts from a total of $N^e$, which can be formulated as:
\begin{equation}
    \mathbf{s}_{\tau}^{c}=\sum_{i=1}^{N^e}R(\mathbf{z}_{\tau}^{c},\mathbf{z}_{\tau}^{\text{task}})_i E_i(\mathbf{\mathbf{z}_{\tau}^{c}}),
\label{eq:moe}
\end{equation}
where $E_i(\cdot)$ stands for the feature representations produced from the $i$-th expert network (i.e., FFN).
The router $R(\cdot,\cdot)$ concurrently receives implicit task directives and modality information, selecting $k$ experts for the current token:
\begin{equation}
    R(\mathbf{z}_{\tau}^{c},\mathbf{z}_{\tau}^{\text{task}})=\text{Softmax}(\text{TopK}(\mathbf{z}_{\tau}^{c}\mathbf{W}^R_1+\mathbf{z}_{\tau}^{\text{task}}\mathbf{W}^R_2,k)),
\end{equation}
\begin{equation}
    \text{TopK}(\mathbf{v},k)_i=\left\{\begin{aligned}
        &\mathbf{v}_i, \quad \text{if } \mathbf{v}_i \text{ is within the top-k} \text{ elements of } \mathbf{v}  \\
        &-\infty, \quad \text{otherwise}
\end{aligned},\right.
\end{equation}
where $\mathbf{W}^R_1\in \mathbb{R}^{d\times N^e}$ and $\mathbf{W}^R_2\in \mathbb{R}^{d\times N^e}$ are learnable projectors for the modality combination token and task token, respectively.
To mitigate the adverse impacts of imbalanced loading, we have incorporated regularization terms aimed at balancing the distribution of expert assignments, following the design and default hyperparameters in previous work \cite{shazeer2017outrageously}.

\subsubsection{\textbf{Patient-level Representation Learning}}
Currently, the representations for various modality combinations under a specific task have been obtained. The next step involves aggregating them into a final patient-level representation.
We design an attention-based mechanism that leverages implicit task information for guidance, achieving task-specific differentiated attention to various modal information. The patient-level representation $\mathbf{s}_{\tau}^p$ is defined as:
\begin{equation}
    \mathbf{s}_{\tau}^p = [\mathbf{z}_{\tau}^\text{task}||\text{LN}(\sum_{c\in \mathcal{C}}\alpha^c \mathbf{s_{\tau}^c})],\quad
    \alpha^c = \frac{\text{exp}(\widetilde{\alpha}^c/\varepsilon)}{\sum_{c\in\mathcal{C}}\text{exp}(\widetilde{\alpha}^c/\varepsilon)},
\label{eq:fusion}
\end{equation}
\begin{equation}
    \widetilde{\alpha}^c = \text{Tanh}([\mathbf{z}_{\tau}^\text{task}||\mathbf{s}_{\tau}^c]\mathbf{W}_1^A)\cdot\mathbf{W}_2^A,
\end{equation}
where $\alpha^c$ is the attention score, $\mathbf{W}_1^A\in\mathbb{R}^{2d\times d}$ and $\mathbf{W}_2^A\in\mathbb{R}^{d\times 1}$ are the weight matrix. 
$\varepsilon$ is a temperature coefficient and $[\cdot||\cdot]$ represents the concatenation operation.

\subsection{Task-specific Prediction Heads}
For different tasks, we employ task-specific prediction heads to obtain the prediction results: $\hat{y}_{\tau} = P_{\tau}(\mathbf{s}_{\tau}^p)$,
where $P_{\tau}(\cdot)$ is the prediction head for the $\tau$-th task. For the binary classification task and multi-label classification task, $P_{\tau}(\cdot)$ contains a linear transformation with a Sigmoid activation. For the multi-class classification task, $P_{\tau}(\cdot)$ contains a liner transformation with a Softmax activation.

The overall objective function for the $\tau$-th task is as follows:
\begin{equation}
    \mathcal{L}_{\tau}=\mathcal{L}^{pred}_{\tau} + \beta \mathcal{L}_{cov},\quad
    \mathcal{L}^{pred}_{\tau} = \frac{1}{N_{\tau}}\sum_{n=1}^{N_{\tau}}\ell_{\tau}(y_{\tau}^{(n)},\hat{y}_{\tau}^{(n)})
\label{eq:loss}
\end{equation}
where $\mathcal{L}_{\tau}^{pred}$ is the prediction loss for the $\tau$-th task, $\ell_{\tau}(\cdot,\cdot)$ denotes a task-specific loss function (binary cross-entropy loss or cross-entropy loss), and $\beta$ is the hyperparameter to strike a balance between the different loss functions.

For ease of understanding, Algorithm \ref{alg} delineates the training procedure of our model.

\begin{algorithm}[t]\small
\SetAlgoLined
\KwInput{\\Multimodal multitask EHR dataset $\{\mathcal{D}_{\tau}\}_{\tau=1}^T$ }
\KwTraining{}
 Initialize weights\;
 \For{epoch in $1,2,...,epoch_{max}$}{
 \For{\text{task} $\tau$ in $1,2,...,T$}{
 \For{B in mini-batches}{
 Extract $\mathbf{H}_{\tau}^m$ of each modality $m$ via Eq.\ref{eq:unimodal}\;
 Obtain the multimodal sequence: $\mathbf{H}_{\tau}^{0}=[\mathbf{h}_{\tau}^{\text{task}},\mathbf{H}_{\tau}^{\text{comb}},\mathbf{H}_{\tau}^{t},\mathbf{H}_{\tau}^{i},\mathbf{H}_{\tau}^{n}]$\;
 Obtain $\mathbf{z}_{\tau}^\text{task}$ and $\mathbf{Z}_{\tau}^\text{comb}$ via Eq.\ref{eq:encoder}-\ref{eq:mhsa}\;
 Obtain the refined representation $\mathbf{s}_{\tau}^c$ for the modality combination $c$ for the specific task via Eq.\ref{eq:moe}\;
 Obtain the patient-level representation $\mathbf{s}_{\tau}^p$ via Eq.\ref{eq:fusion}\;
 Make prediction for the task $\tau$\;
 Update the parameters by optimizing Eq.\ref{eq:loss}.
 }
 }
 }
 \caption{Algorithm of FlexCare}\label{alg}
\end{algorithm}

\begin{table}[t]
  \caption{Statistics of the datasets for multiple tasks.}
  \label{tab:stat}
  \begin{tabular}{lcp{1.6cm}<{\centering}p{1.6cm}<{\centering}p{1.6cm}<{\centering}}
    \toprule
    \multirow{2}{*}{Task} & \multirow{2}{*}{\# Number} & \multicolumn{3}{c}{Missing rate per modality} \\
    & & Time-series & Image & Note \\
    \midrule
    IHM & 26,318 & 0\% & 76.40\% & 7.49\% \\
    LOS  & 59,495 & 0\% & 85.16\% & 8.27\%  \\
    DEC  & 59,269 & 0\% & 85.15\% & 8.24\%  \\
    PHE     & 59,798 & 0\% & 81.94\% & 8.30\%  \\
    REA     & 55,712 & 0\% & 82.38\% & 8.06\%  \\
    DIA       & 132,576 & 76.34\% & 0\% & 32.56\%  \\

    \bottomrule
  \end{tabular}
\end{table}

\begin{table*}[t]\renewcommand\arraystretch{1.1}
  \caption{Performance comparison between baselines and the proposed method on multiple tasks. The best and second-best results are highlighted in bold and underline, respectively.}
  \label{tab:main}
  \begin{tabular}{ll|p{1.7cm}<{\centering}p{1.7cm}<{\centering}p{1.7cm}<{\centering}p{1.7cm}<{\centering}p{2cm}<{\centering}|p{1.7cm}<{\centering}p{1.7cm}<{\centering}}
    \toprule
    Task & Metric & MedFuse~\cite{hayat2022medfuse} & MT~\cite{ma2022multimodal} & M3Care~\cite{zhang2022m3care} & MMF~\cite{lee2023learning} & MultiModN~\cite{swamy2023multimodn} & FlexCare-st & FlexCare\\
    \midrule
    \multirow{2}{*}{IHM} & AUROC & 0.8772 (0.003) & 0.8726 (0.002) & 0.8732 (0.006) & \underline{0.8804 (0.001)} & 0.8751 (0.002) & 0.8749 (0.004) & \pmb{0.8823 (0.002)}\\
    & AUPRC & \underline{0.5158 (0.006)} & 0.5133 (0.008) & 0.5148 (0.017) & 0.5136 (0.010) & 0.5055 (0.006) & 0.5116 (0.007) & \pmb{0.5372 (0.006)}\\
    \midrule
    \multirow{2}{*}{LOS}  & ma-F1 & 0.1487 (0.006) & 0.1531 (0.006) & \underline{0.1549 (0.007)} & \pmb{0.1554 (0.006)} & 0.1503 (0.010) & 0.1492 (0.005) & 0.1479 (0.005)\\
    & mi-F1 & 0.6289 (0.005) & 0.6298 (0.003) & 0.6267 (0.004) & 0.6282 (0.004) & 0.6307 (0.006) & \underline{0.6317 (0.001)}& \pmb{0.6358 (0.003)}\\
    \midrule
    \multirow{2}{*}{DEC} & AUROC & 0.9396 (0.002) & 0.9409 (0.001) & 0.9406 (0.004) & 0.9435 (0.001) & \underline{0.9470 (0.001)} & 0.9420 (0.002)& \pmb{0.9538 (0.001)}\\
    & AUPRC & 0.4782 (0.006) & 0.4792 (0.010) & 0.4911 (0.011) & \underline{0.4981 (0.008)} & 0.4922 (0.005) & 0.4926 (0.010) & \pmb{0.5123 (0.006)}\\
    \midrule
    \multirow{2}{*}{PHE} & ma-AUROC & 0.8340 (0.001) & 0.8362 (0.001) & \underline{0.8429 (0.001)} & \pmb{0.8446 (0.001)} & 0.8424 (0.000) & 0.8417 (0.000)& 0.8393 (0.005)\\
    & mi-AUROC & 0.8785 (0.001) & 0.8769 (0.001) & \underline{0.8830 (0.001)} & \pmb{0.8845 (0.000)} & 0.8826 (0.000) & 0.8820 (0.000)& 0.8803 (0.004)\\
    \midrule
    \multirow{2}{*}{REA} & AUROC & 0.7598 (0.002) & 0.7585 (0.002) & 0.7618 (0.001) & \underline{0.7627 (0.002)} & 0.7622 (0.001) & 0.7604 (0.002) & \pmb{0.7680 (0.002)}\\
    & AUPRC & \underline{0.3618 (0.003)} & 0.3481 (0.008) & 0.3562 (0.003) & 0.3482 (0.006) & 0.3526 (0.004) & 0.3517 (0.003) & \pmb{0.3702 (0.004)}\\
    \midrule
    \multirow{2}{*}{DIA} & ma-AUROC & 0.6651 (0.007) & 0.6715 (0.005) & \underline{0.6756 (0.006)} & 0.6692 (0.002) & 0.6717 (0.005) & 0.6750 (0.005)& \pmb{0.6845 (0.006)}\\
    & mi-AUROC & 0.8920 (0.002) & 0.8960 (0.002) & 0.8955 (0.001) & \underline{0.8960 (0.001)} & 0.8944 (0.001) & 0.8948 (0.001)& \pmb{0.8984 (0.001)}\\
    
    \bottomrule
  \end{tabular}
\end{table*}

\section{Experiment}
In this section, we evaluate our proposed FlexCare model focusing on the following research questions:
\begin{itemize}[leftmargin=*,noitemsep,topsep=2pt]
    \item RQ1: How does FlexCare perform on multiple real-world tasks?
    \item RQ2: How does each component influence the performance of FlexCare?
    \item RQ3: Whether FlexCare leverages the cross-task synergy?
    \item RQ4: How is the extensibility of FlexCare?
\end{itemize}

\subsection{Experimental Settings}
\subsubsection{\textbf{Datesets}} 
In this study, we use three EHR datasets: MIMIC-IV\footnote{https://physionet.org/content/mimiciv/2.0/}, MIMIC-CXR JPG\footnote{https://physionet.org/content/mimic-cxr-jpg/2.0.0/} and MIMIC-IV-NOTE\footnote{https://physionet.org/content/mimic-iv-note/2.2/}~\cite{johnson2020mimic,johnson2019mimic,johnson2023mimic}.
Since these datasets share the same patient cohort, we gathered time-series data from MIMIC-IV, X-ray images from MIMIC-CXR, and clinical notes from MIMIC-NOTE, thereby constructing a comprehensive multimodal patient dataset.
We divide the patients into a training set, a validation set, and a test set in a 7:1:2 ratio.
On the basis of the multimodal dataset, we extract sub-datasets for 6 tasks: in-hospital-mortality (IHM), length-of-stay (LOS), decompensation (DEC), phenotyping (PHE), readmission (REA) and diagnosis (DIA). It is imperative to note that each of these tasks utilizes heterogeneous datasets, meaning the input data and output labels vary across tasks.
Table \ref{tab:stat} provides detailed statistics of datasets for multiple tasks, revealing variations in sample sizes and modality messiness among different tasks.

\subsubsection{\textbf{Baselines and Implementation Details}} 
To assess the effectiveness of our proposed model, we compare it against the following multimodal baselines: 
MedFuse~\cite{hayat2022medfuse}, MT~\cite{ma2022multimodal}, M3Care~\cite{zhang2022m3care}, MMF~\cite{lee2023learning} and MultiModN~\cite{swamy2023multimodn}.

For binary classification tasks (IHM, DEC, and REA), we use AUROC and AUPRC as evaluation metrics. For multi-class tasks (PHE and DIA), we use macro-AUROC and micro-AUROC; and for multi-label classification task (LOS), we use macro-F1 and micro-F1.

Due to differences in the problem setting, previous multitask models are not applicable to our constructed dataset. Because these datasets are independent, each sample has a label for only one task.
Moreover, it should be noted that as some of the above models do not encompass all modalities present in our dataset, we have extended them accordingly while ensuring the consistency of the foundational embedding layer across all models (i.e., a linear projection, patch projection and the pre-trained \& frozen BioBERT).
Given the varying training difficulties across different tasks, 
we employ distinct weights for each task in FlexCare and mitigate the issue of inconsistent convergence rates among multitask learning through weight decay.

\subsection{Model Performance (RQ1)}
In the experiments, baseline models are independently trained for each task, whereas FlexCare is trained on all tasks within a single model.
As shown in Table~\ref{tab:main}, two key conclusions can be drawn: \textbf{(1) Compared to single-task baseline models, FlexCare achieves competitive results across various evaluation metrics while also being adaptive to a range of tasks.} 
Overall, the performance differences among the baseline models are not significant. This could be attributed to the fact that after replacing with the unified unimodal encoders, various multimodal fusion strategies under a single-task are capable of effectively addressing the current issue, yet they have reached a performance plateau.
Unlike other models that are optimized for a single task, our multi-task model aims at the overall performance improvement and may not achieve optimal results in some individual metrics across all 6 tasks.
By means of the synergistic effect between tasks, our model achieves the best results on the IHM, DEC, REA, and DIA tasks, with performance on the remaining two tasks closely matching that of the baselines.
\textbf{(2) Through leveraging cross-task synergy, tasks that are closely related can benefit from each other.}
FlexCare achieves significant performance improvements on the IHM and DEC tasks, which are related to patient mortality risk.
This demonstrates that FlexCare can leverage cross-task synergy to acquire additional knowledge, aiding in achieving superior performance for related tasks. 
Moreover, comparing to FlexCare-st that is trained independently for each task, we can observe that the strategy of heterogeneous multitask joint training is both rational and effective, with certain tasks benefiting from other related tasks.

\begin{figure*}[tbp]\setlength{\belowcaptionskip}{-8pt}
  \centering
  \includegraphics[width=0.95\linewidth]{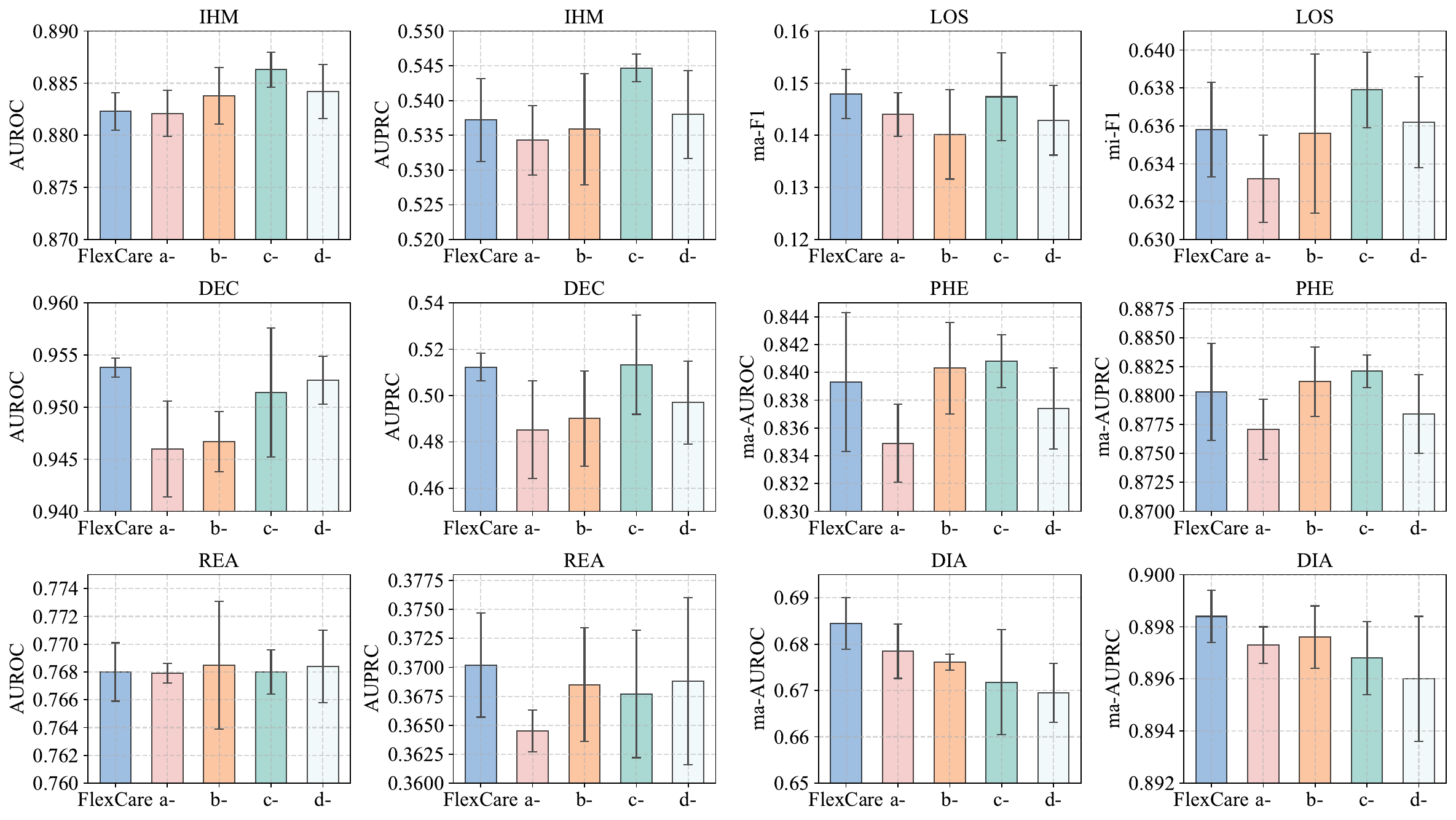}
  \caption{Results of ablation study.}
  \label{fig:ablation}
\end{figure*}

\begin{table}[t]
  \caption{Ablation study setup.}
  \label{tab:abla}
  \begin{tabular}{lp{2cm}p{1.8cm}p{1.8cm}}
    \toprule
    Model & Modality Combination tokens & Representation Decorrelation & Task/Modality-aware MoE \\
    \midrule
    FlexCare$_\text{a-}$ & $\times$ & $\times$ & $\times$ \\
    FlexCare$_\text{b-}$ & \checkmark & $\times$ & $\times$ \\
    FlexCare$_\text{c-}$ & \checkmark & \checkmark & $\times$ \\
    FlexCare$_\text{d-}$ & \checkmark & $\times$  & \checkmark \\
    \midrule
    FlexCare & \checkmark & \checkmark & \checkmark \\
    \bottomrule
  \end{tabular}
\end{table}

\subsection{Analysis of Model Design (RQ2)}
\subsubsection{\textbf{Ablation Study}}
To evaluate the contributions of the different modules of our model to the prediction performance, we conduct an ablation study by comparing three variants of the model: FlexCare$_\text{a-}$, FlexCare$_\text{b-}$, FlexCare$_\text{c-}$ and FlexCare$_\text{d-}$. The specific setting is outlined in Table \ref{tab:abla}, where different models comprise various combinations of three modules.

As shown in Figure \ref{fig:ablation}, the performance disparity between FlexCare and both FlexCare$_\text{a-}$ and FlexCare$_\text{b-}$ highlights the effectiveness of modality combination tokens and covariance regularization.
Furthermore, FlexCare$_\text{c-}$ performs well on several tasks, but worse on others (e.g., DEC, REA, DIA), even falling below that of FlexCare$_\text{a-}$ and FlexCare$_\text{b-}$. Additionally, FlexCare$_\text{c-}$ exhibits a higher overall standard deviation, indicating less stability. These results corroborate the phenomenon of negative interface in multitask learning mentioned above.

\begin{figure}[t]
  \centering
  \includegraphics[width=1\linewidth]{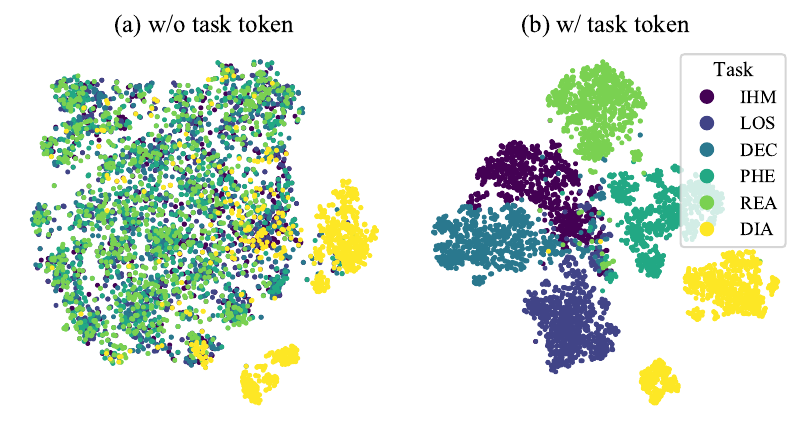}
  \caption{Visualization of patient-level representation learned w/o and w/ the task token.}
  \label{fig:task_emb}
\end{figure}

\subsubsection{\textbf{Analysis of Learnable Task Token}}
To facilitate multimodal representation learning for specific tasks, we incorporate task information as input and learn a corresponding task token for each task.
To demonstrate the impact of the task token on the final representations, we randomly select 30 mini-batch samples (960 samples in total) for each task, comparing scenarios with and without the use of the task token.
Figure \ref{fig:task_emb} shows the visualized results of final embeddings learned w/o and w/ the task token.
It is observed that without explicitly providing task information as input to the model, the representations of samples from different tasks are intermingled and indistinguishable. However, upon inputting task information, the representations of various tasks are effectively segregated. Moreover, from Figure \ref{fig:task_emb}(b), it should be noted that samples from the five tasks primarily based on time-series modalities are more clustered together, whereas samples from the DIA task, which relies on the imaging modality, are situated farther from the others. This indicates the ability of FlexCare to learn the modality-specific differences inherent to each task.

\subsection{Analysis of Cross-Task Synergy (RQ3)}
\subsubsection{\textbf{Expert Selection}}
\begin{figure}[tbp]
  \centering
  \includegraphics[width=1\linewidth]{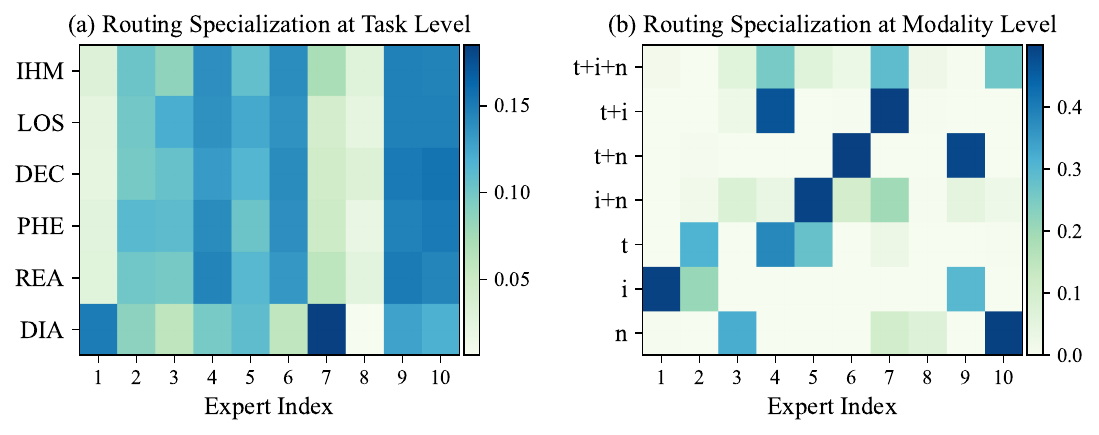}
  \caption{Routing specialization at task and modality levels.}
  \label{fig:expert}
\end{figure}

As shown in Figure \ref{fig:expert}, we visualize the frequency of experts being selected for each task and each modality combination.
From Figure \ref{fig:expert}(a), we observe that the experts employed for task DIA exhibit significant differences compared to those used for other tasks. This discrepancy arises because DIA primarily relies on the imaging modality, with other modalities serving as auxiliary information, whereas other tasks predominantly depend on time-series modality.
Figure \ref{fig:expert}(b) presents the results of modality-level routing specialization, reflecting the distinct capabilities of each expert, which aids in the refined processing of multimodal information.

\subsubsection{\textbf{Interaction Between Tasks}}
\begin{figure}[tbp]
  \centering
  \includegraphics[width=1\linewidth]{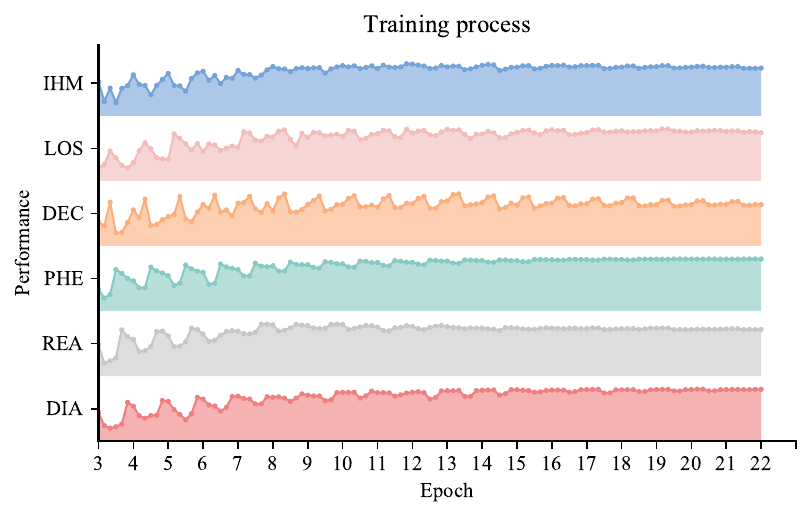}
  \caption{Performance analysis of each task in the training process.}
  \label{fig:training}
\end{figure}

To demonstrate whether FlexCare effectively utilizes cross-task synergy, we analyze the impact of the current training task on other tasks.
Specifically, we sequentially train the six tasks in each epoch: IHM, LOS, DEC, PHE, REA, and DIA.
Upon the completion of training for the current task, we report the test performance across all tasks.
As illustrated in Figure \ref{fig:training}, it can be observed that after the training of a specific task, the performance of other tasks does not necessarily decline; rather, it may even exhibit improvement.
For instance, the performance of tasks IHM and DEC maintains an upward trend even when other tasks are being trained.
This demonstrates the ability of FlexCare to effectively leverage cross-task synergy, enhancing the performance of a task by other related tasks.

\subsection{Analysis of Extensibility (RQ4)} 
As a flexible multimodal multitask model, FlexCare can be adapted to other new tasks. We construct a new task, Diagnosis-related Group (DRG) prediction~\cite{liu2021early}, which primarily relies on clinical notes. This task features a sizable dataset, encompassing 236,770 samples.  We conduct experiments under two settings, training for 40 epochs with 1$\%$ of the training data and 100$\%$ of the training data, respectively.
Table~\ref{tab:drg} presents the experimental results, where FlexCare$_\text{pretrain}$ employs the model trained on the six tasks mentioned previously, while FlexCare denotes the model trained from scratch on the current task.
It can be observed that with limited training data, both the baseline models and FlexCare that trained from scratch, do not perform as well as pre-trained FlexCare. However, when utilizing 100$\%$ of the training data, the advantage of pre-trained FlexCare is diminished due to the significantly larger sample size of this task compared to the others. 
This experiment demonstrates FlexCare can be flexibly extended to new tasks, achieving commendable performance even with smaller data sizes.

\begin{table}[tb]\renewcommand\arraystretch{1.1}
  \caption{Results of DRG prediction.}
  \label{tab:drg}
  \begin{tabular}{lp{1.2cm}<{\centering}p{1.2cm}<{\centering}p{1.2cm}<{\centering}p{1.2cm}<{\centering}}
    \toprule
    \multirow{2}{*}{Method} & \multicolumn{2}{c}{1\% Labeled} & \multicolumn{2}{c}{100\% Labeled} \\
    & ma-F1 & mi-F1 & ma-F1 & mi-F1 \\
    \midrule
    MedFuse & 0.0054 & 0.1125 & 0.2149 & 0.4367 \\
    MT & 0.0128 & 0.1567 & 0.2283 & 0.4322 \\
    M3Care & 0.0320 & 0.2491 & 0.2424 & 0.4402 \\
    MMF & 0.0253 & 0.2316 & 0.2474 & 0.4423 \\
    MultiModN & 0.0228 & 0.2162 & 0.2424 & 0.4412 \\
    \hline
    FlexCare$_\text{pretrain}$ & \pmb{0.0396} & \pmb{0.2695} & 0.2404 & 0.4387 \\
    FlexCare & 0.0294 & 0.2390 & \pmb{0.2487} & \pmb{0.4440} \\
    \bottomrule
  \end{tabular}
\end{table}

\section{Conclusion}
In this paper, we introduce FlexCare, a unified healthcare prediction model that flexibly accommodates incomplete multimodal inputs and adapts to multiple tasks. 
It is endowed with multitasking capabilities realized through asynchronous multiple single-task predictions.
Specifically, the task-agnostic multimodal information extraction module is designed to thoroughly capture information across a spectrum of modality combination patterns. 
Meanwhile, a token-level covariance regularization method is developed to prevent different modality combination tokens from encoding similar information.
Furthermore, we propose the task-guided hierarchical multimodal fusion module to learn adaptive representation tailored to the specific task.
In addition, we experimented on multiple tasks from MIMIC-IV/MIMIC-CXR/MIMIC-NOTE datasets to show the effectiveness of the proposed model.

In future work, we will investigate effective solutions 
to address the issues of gradient conflicts and inconsistencies in convergence rates during multitask training, contributing to the advancement of a general prediction model in the healthcare domain.

\section*{ACKNOWLEDGMENTS}
This work was supported in part by the National Key Research and Development Program of China under Grant No.2021ZD0140407, the Beijing Natural Science Foundation under Grant No.7222313, and the National High Level Hospital Clinical Research Funding under Grant No.2022-PUMCH-C-041.

\bibliographystyle{ACM-Reference-Format}
\bibliography{sample-base}

\appendix

\section{DETAILS OF EXPERIMENTAL SETTINGS}
\subsection{Tasks}

Following previous works for dataset creation, we extract sub-datasets for 7 tasks from MIMIC-IV dataset, 6 of which are utilized for multitask performance comparison, and the remaining one for extensibility analysis:
\begin{itemize}[leftmargin=*,noitemsep,topsep=2pt]
    \item \textbf{In-hospital mortality (IHM)}: predicts in-hospital mortality based on the first 48 hours of an ICU stay. This is a binary classification task.
    \item \textbf{Length-of-stay (LOS)}: predicts remaining time spent in ICU. We define it as a classification problem with 10 classes (one for ICU stays shorter than a day, seven day-long buckets for each day of the first week, one for stays of over one week but less than two, and one for stays of over two weeks).
    \item \textbf{Decompensation (DEC)}: predicts whether a patient will decease in the next 24 hours. This is a binary classification task.
    \item \textbf{Phenotyping (PHE)}: classifies which of 25 acute care conditions are present in a given patient ICU stay record. This is a multilabel classification problem.
    \item \textbf{Readmission (REA)}: predicts whether the patient will be readmitted within the next 30 days. This is a binary classification task.
    \item \textbf{Diagnosis (DIA)}: predicts diagnosis based on chest radiograph. This is a multilabel classification problem.
    \item \textbf{Diagnosis-related Group (DRG)}: predicts the DRG code assigned to the patient. This is a multi-class classification task.
\end{itemize}

\subsection{Data Preprocessing}
\begin{itemize}[leftmargin=*,noitemsep,topsep=2pt]
\item \textbf{Time series data}: 17 variables are sampled every one hour, and the missing value is imputed using the previous one. Then, they are discretized and standardized to obtain the 76-dimension input at each time-step.
\item \textbf{Image data}: Images classified as AP (Anteroposterior) are retained. Following the previous work \cite{hayat2022medfuse}, during the training phase, the images are resized to 256 $\times$ 256 pixels, and random horizontal flip and random affine transformation are applied, followed by cropping to 224 $\times$ 224 pixels. For validation and testing phases, only resizing and center cropping to 224 $\times$ 224 pixels are performed.
\item \textbf{Note data}: We only extract the Past Medical History of the clinical notes for tasks other than DRG to avoid information leakage. For DRG task, we extract Brief Hospital Course part as the overall admission summary.
\end{itemize}

\subsection{Baselines}
The detailed description about baselines is as follows:
\begin{itemize}[leftmargin=*,noitemsep,topsep=2pt]
    \item \textbf{MedFuse} \cite{hayat2022medfuse}: develops an LSTM-based fusion module that can accommodate uni-modal as well as multi-modal input.
    \item \textbf{MT} \cite{ma2022multimodal}: employs the Transformer architecture with attention mask to fuse multimodal data with missing modality.
    \item \textbf{M3Care} \cite{zhang2022m3care}: imputes the information of the missing modalities in the latent space from the similar neighbors of each patient. 
    \item \textbf{MMF} \cite{lee2023learning}: learns the EHR data with missing modal by the modality-aware attention with skip bottleneck.
    \item \textbf{MultiModN} \cite{swamy2023multimodn}: sequentially inputs any number or combination of modalities into a sequence of modality-specific encoders and can skip over missing modalities.
\end{itemize}

\subsection{Model Implementation}
Our model is implemented using PyTorch 1.10.0 and Python 3.7.9.
The experiment environment is a machine equipped with Ubuntu 20.04 and NVIDIA GeForce RTX 3090 GPU.
For models with transformer encoders, the number of transformer layers is set to 4 and the number of attention heads is set to 2. For all models, the dimension of the hidden layer is 128.
The number of the selected experts $k$ and the total experts $N^e$ are 2 and 10.
Due to large search space for 6 task weights, we set them with \{0.2,0.5,0.2,1,0.2,0.2\} based on loss magnitude and training difficulty of single-task models.
We use Adam as the optimizer with batch size 32, learning rate 1e-3 or 5e-4.
We train baseline models for 40 epochs and modify random seeds for five repetitions.

\section{Further analysis on Task token}
In Section 5.3.2, we visualize the patient-level representation learned w/o and w/ the task token.
Here, we present quantitative experimental results under two scenarios. As observed in Table \ref{tab:task}, the performance across various tasks declines when task information is not provided to the model, which is precisely because the representations of different tasks are mixed together and cannot be clearly distinguished.

\begin{table}[b]\renewcommand\arraystretch{1.1}
  \caption{Performance comparison using FlexCare w/o and w/ the task token.}
  \label{tab:task}
  \begin{tabular}{ll|cc}
    \toprule
    Task & Metric & w/o task token & FlexCare\\
    \midrule
    \multirow{2}{*}{IHM} & AUROC & \pmb{0.8831 (0.003)} & 0.8823 (0.002) \\
    & AUPRC & 0.5292 (0.007) & \pmb{0.5372 (0.006)} \\
    \midrule
    \multirow{2}{*}{LOS}  & ma-F1 & 0.1321 (0.004) & \pmb{0.1479 (0.005)} \\
    & mi-F1 & 0.6351 (0.004) & \pmb{0.6358 (0.003)} \\
    \midrule
    \multirow{2}{*}{DEC} & AUROC & 0.9463 (0.001) & \pmb{0.9538 (0.001)}\\
    & AUPRC & 0.4891 (0.010) & \pmb{0.5123 (0.006)}\\
    \midrule
    \multirow{2}{*}{PHE} & ma-AUROC & 0.8338 (0.005) & \pmb{0.8393 (0.005)}\\
    & mi-AUROC & 0.8747 (0.005) & \pmb{0.8803 (0.004)} \\
    \midrule
    \multirow{2}{*}{REA} & AUROC & \pmb{0.7698 (0.002)} & 0.7680 (0.002)\\
    & AUPRC & 0.3625 (0.007) & \pmb{0.3702 (0.004)}\\
    \midrule
    \multirow{2}{*}{DIA} & ma-AUROC & 0.6778 (0.001) & \pmb{0.6845 (0.006)}\\
    & mi-AUROC & 0.8974 (0.001) & \pmb{0.8984 (0.001)} \\
    
    \bottomrule
  \end{tabular}
\end{table}

\end{document}